Field-testing items using artificial intelligence: Natural language processing with transformers

Hotaka Maeda, PhD, Smarter Balanced

hotaka.maeda@smarterbalanced.org

Abstract

Five thousand variations of the RoBERTa model, an artificially intelligent "transformer" that can understand text language, completed an English literacy exam with 29 multiple-choice questions. Data were used to calculate the psychometric properties of the items, which showed some degree of agreement to those obtained from human examinee data.

## Introduction

Field-testing is costly and time-consuming (Jiao & Lissitz, 2020). There has been a variety of efforts to limit the need for extensive field-testing of new items (e.g., Glas & van der Linden, 2003). Some have turned to natural language processing (NLP) to approximate item difficulty and discrimination from the item text (Benedetto et al., 2020; Laverghetta et al., 2021; Luger, 2016). NLP is a branch of artificial intelligence (AI) concerned with providing computers an understanding of text and language. Currently, the field of NLP is led by the state-of-the-art class of deep learning model architecture called the transformer (Vaswani et al., 2017). The core of transformers is the multiheaded attention mechanism, which create the meaning of each word efficiently by identifying its contextual relationship with the other words. For example, transformers are able to distinguish the difference in the meaning of "check" in the phrases "write a check" and "check the engine". Transformers also excel at understanding the meaning of relatively long text.

One recently introduced transformers is RoBERTa (Yinhan et al., 2019), which is based on the BERT model (Devlin et al., 2018) and has been pre-trained using 160GB of English text. RoBERTa is able to answer both open-ended and multiple-choice questions (MCQ) by selecting phrases that have the highest probability of matching the context. There have not been attempts to use transformers to generate human-like item response data for use in psychometric analyses. In this proof-of-concept study, RoBERTa with varying levels of intelligence was created by manipulating its vocabulary, and used to generate item response data for English MCQ items in an attempt to estimate their item parameters. These RoBERTa-based item parameters were compared with those from human data.

## Method

### RoBERTa

The 1.4GB RoBERTa-large model (Liu et al., 2019) fine-tuned using the RACE dataset (Lai et al., 2017) was used. RACE is a publicly available English reading comprehension dataset with 27,933 passages and 97,867 four-option MCQ items, which is a design similar to the items included in the current study. RoBERTa answered 85.2% of the RACE items correctly.

RoBERTa holds a vocabulary of 50,265 tokens (i.e., words, sub-words, or punctuation), each with 1,024 numerical weights called "word embeddings" that define its latent meaning (Turian et al., 2010). In each iteration of this study, a random proportion $U(0,1)$ of the 50,265 tokens was randomly selected, then their weights were set to zero. This manipulation effectively forces RoBERTa to "forget" the meanings of some words. This was a simple, computationally fast, and effective way to create less intelligent variations of RoBERTa.

Item selection

Third-grade English literacy four-option MCQ were used in the study. Items regarding grammar, bolded words, underlined words, fill-in-the-blank, images, or audio were removed. Items with more than 512 tokens were removed, as RoBERTa can process only 512 tokens simultaneously. There were 34 items that qualified, 5 of which RoBERTa answered incorrectly (85% correct) and were removed. Therefore, 29 items were included in the study (see Figure 1).

Data collection

The 2-parameter logistic model (2PL) parameters for the 29 items were previously estimated using human 3$^{rd}$ grade student data in the United States ($N$=814 to 5,283 per item), with ability $\theta \sim N(0,1)$. $N$=5,000 variations of RoBERTa completed the 29-item exam using the transformers 4.18.0 and PyTorch 1.12.1 libraries in Python. RoBERTa provided the probability that each response option may be correct. Based on these probabilities, a response was randomly selected for each item.

Parameter estimation

RoBERTa's item responses were used to estimate the 2PL model with a 1.7 scaling factor using the mirt package in R (Chalmers, 2012). Mean and *SD* of $\theta$ were freely estimated. To place human and RoBERTa parameters on the same scale, item parameters were estimated one item at a time, with all others fixed (i.e., anchored) to the human-based values. RoBERTa's ability estimates were obtained using both human ($\hat{\theta}_H$) and RoBERTa ($\hat{\theta}_R$) item parameters, with maximum a-posteriori with a weak prior $\theta \sim N(0,100)$. Human and RoBERTa statistics were compared using mean bias (RoBERTa minus human), root-mean-squared-error (RMSE), and Spearman correlation.

Result

RoBERTa's mean score was .47 (*SD*=0.23, Cronbach's $\alpha$=.87). A high negative correlation ($r$=-.86) between $\hat{\theta}_R$ and the proportion of word embedding weights set to zero showed that the RoBERTa's intelligence was manipulated effectively (see Figure 2). Positive correlations were found between the corresponding human and RoBERTa-based item statistics ($r$=.39 to .47; see Table 1 and Figure 3). Ability estimates $\hat{\theta}_H$ and $\hat{\theta}_R$ were similar (bias=0.03, RMSE=0.22, $r$=.99), showing little practical difference of using human and RoBERTa-based item parameters on test scores (see Figure 4).

Discussion

This preliminary study demonstrated the potential of approximating the item parameters of new items using transformer NLP models. To the author's knowledge, this was also the first study to purposely and randomly downgrade the AI's intelligence in an attempt to generate human-like item responses. There was some agreement between humans and RoBERTa about which items were difficult or discriminate well. However, there were still considerable disagreements, which may be explained partly that human intelligence is not as simple as a random proportion of known vocabulary. A more complex manipulation of the AI's intelligence may be necessary to mimic the diverse knowledge levels and patterns of the target human population.

Potentially, using AI can bypass or reduce the need to administer field-test items to real humans, which could save resources for testing organizations and reduce the stress on examinees to complete more items. Unlike human examinees, AI could take thousands of items in a large item pool without fatigue or exposure, which may be replicated until ability and item parameter estimation error are near zero.

Limitations included the lack of access to individual human data, and RoBERTa's incompatibility for answering some items. In the future, transformers could be fine-tuned using operational items similarly designed to the field-test items. Ethics of using AI for field-testing must also be considered, such as how to incorporate differential item functioning analysis in the process.

Figure 1. An example 3rd grade English literacy item from the study that RoBERTa answered correctly (Item #10 on Figure 2)

A student is writing a report on cheetahs for her class. The student wants to revise a paragraph from the report to give it a stronger conclusion. Read the draft of a paragraph from the report and complete the task that follows.

Cheetahs are very fast runners. They are the fastest land animals in the world. They can run up to 75 miles per hour! They hunt animals like rabbits and deer. Most cheetahs live in Africa. They are the only big cats that cannot roar. They also cannot climb trees. Cheetahs are smaller than lions and tigers. Their fur is tan with black spots all over. There are only about 10,000 cheetahs left in the world.

Choose the sentence that is the **best** conclusion to the paragraph from the student's report.

(A) Cheetahs are usually lonely animals.

(B) Cheetahs have between 2 and 4 cubs in a litter.

(C) Cheetahs have 2,000 to 3,000 spots on their fur.

(D) Cheetahs are amazing animals that may soon be all gone.

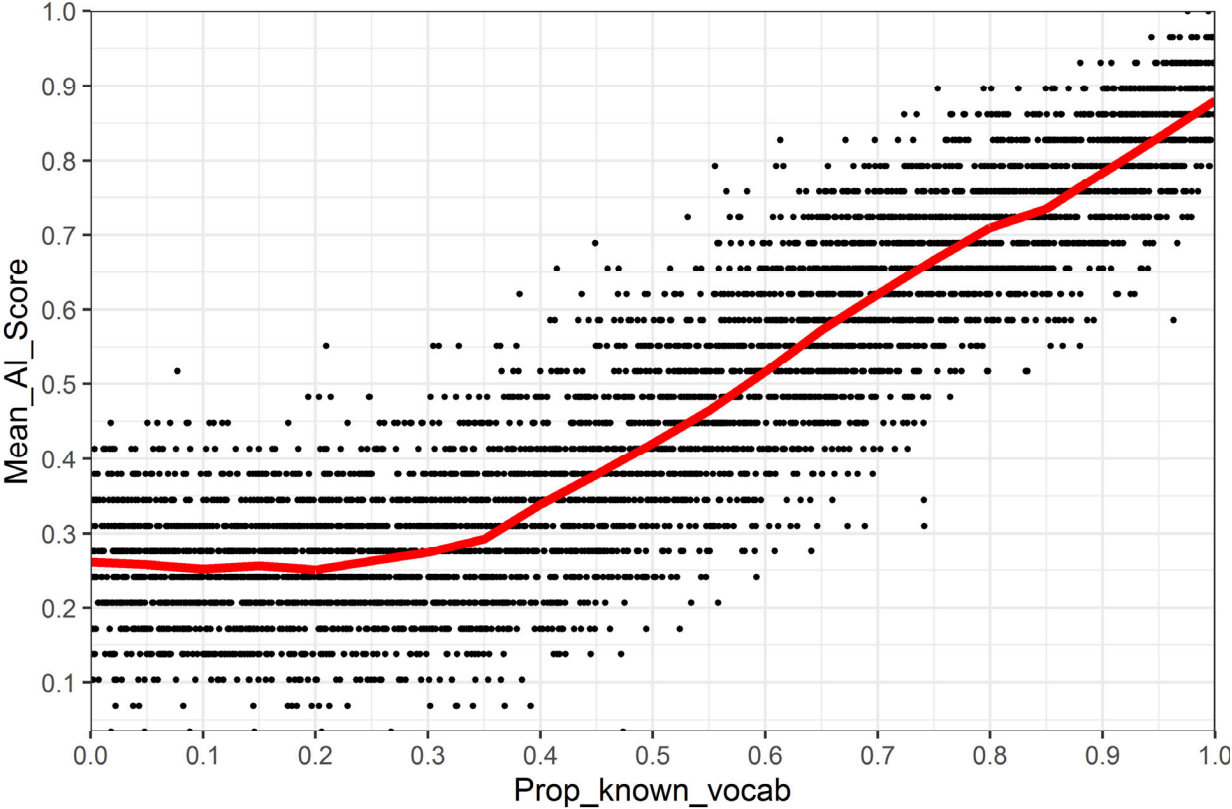

Figure 2. RoBERTa's mean test score by proportions of original vocabulary retained

Figure 3. 2PL item response functions estimated from human and RoBERTa's data (29 items, sorted by item difficulty)

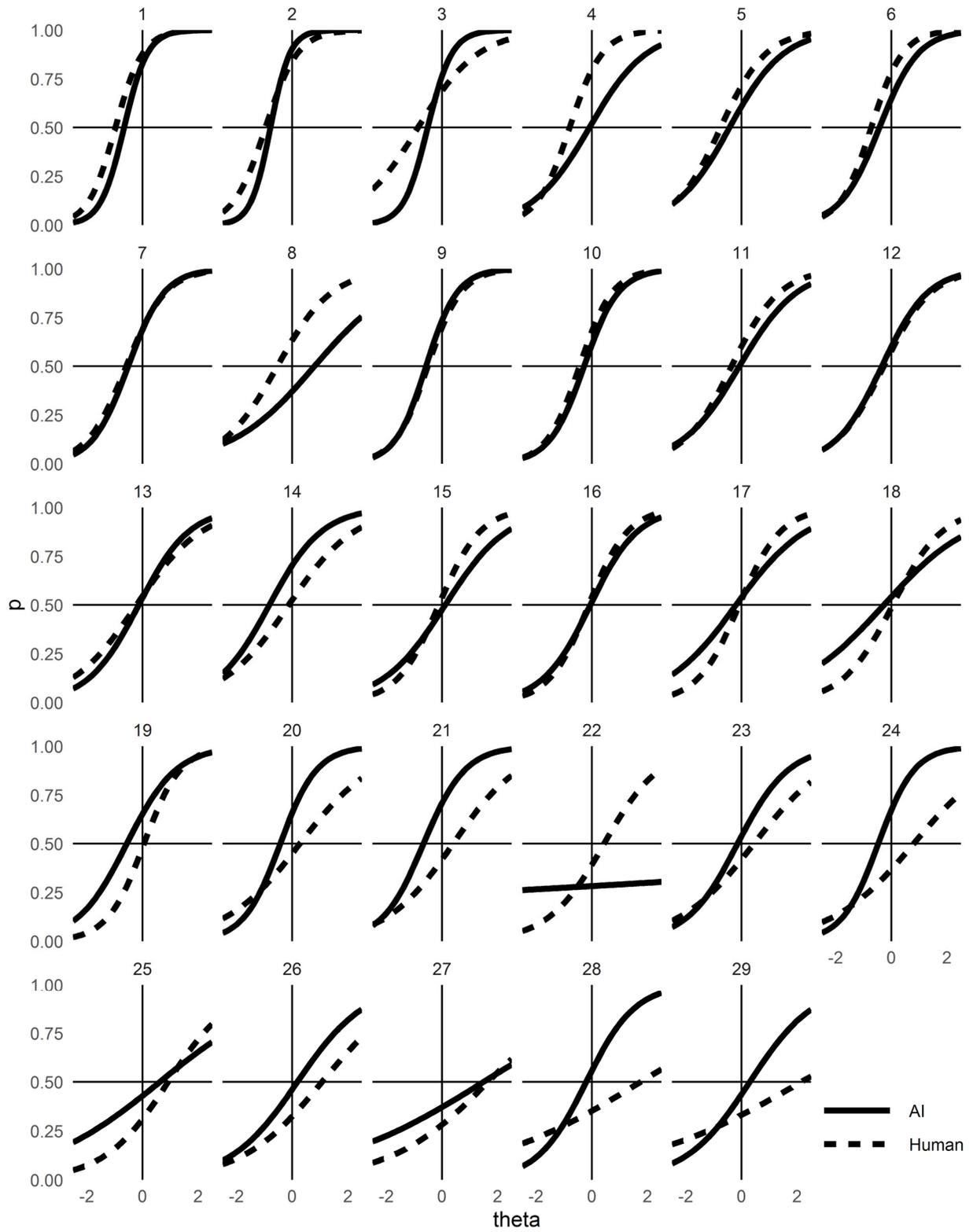

Figure 4. RoBERTa's estimated theta based on item parameters estimated from human and RoBERTa data

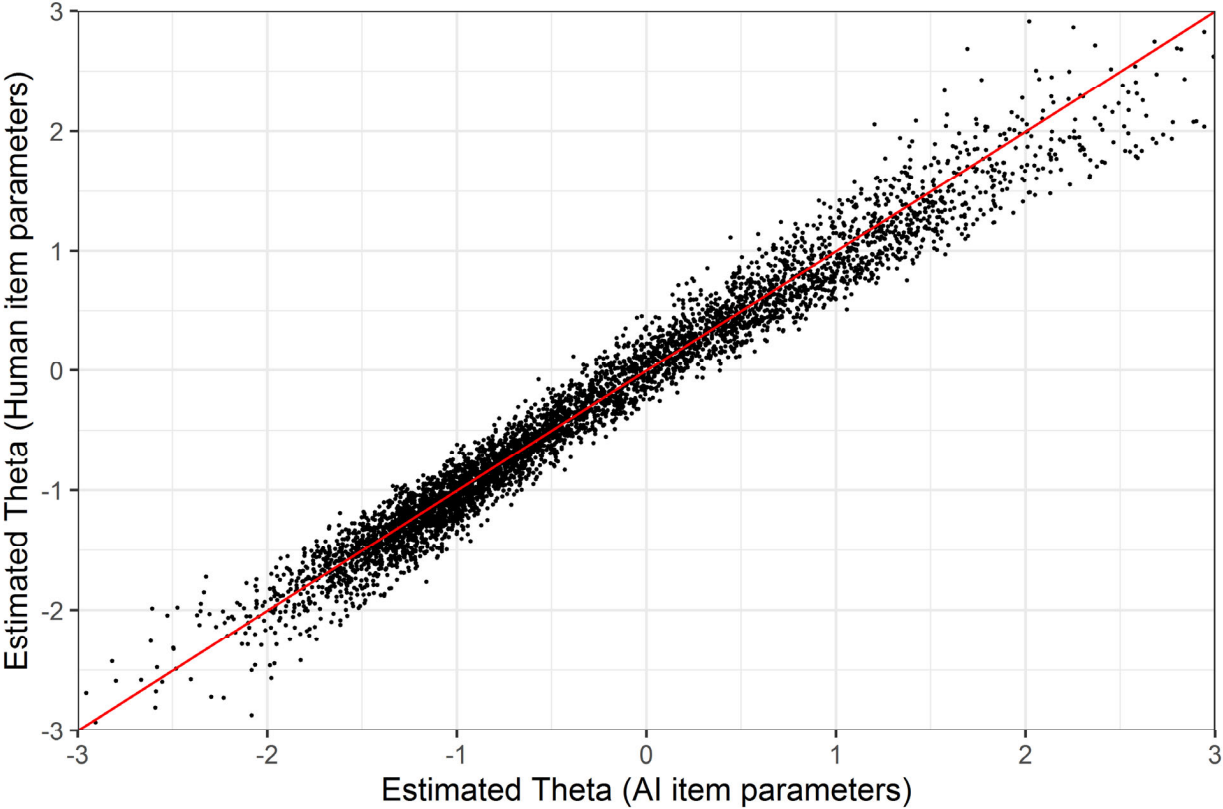

Table 1.

| | Statistic | RoBERTa ability | 2PL discrimination | 2PL difficulty[a] | Proportion correct[b] | Item-total correlation[b] |
|---|---|---|---|---|---|---|
| Human data | | | | | | |
| | Mean | -0.32 | 0.66 | 0.05 | .52 | .50 |
| | SD | 1.08 | 0.26 | 0.83 | .13 | .17 |
| | Median | -0.53 | 0.65 | -0.12 | .52 | .55 |
| | Min | -3.66 | 0.19 | -0.97 | .32 | .16 |
| | Max | 8.62 | 1.2 | 2.14 | .76 | .73 |
| RoBERTa data | | | | | | |
| | Mean | -0.29 | 0.7 | 0.6 | .45 | .47 |
| | SD | 1.12 | 0.36 | 4.25 | .08 | .10 |
| | Median | -0.55 | 0.64 | -0.27 | .45 | .47 |
| | Min | -6.36 | 0.02 | -0.83 | .27 | .20 |
| | Max | 8.15 | 1.68 | 22.55 | .65 | .61 |
| Comparison | | | | | | |
| | Bias | 0.03 | 0.04 | 0.55 | | |
| | RMSE | 0.22 | 0.33 | 4.21 | | |
| | Spearman Correlation | .99 | .39 | .45 | .47 | .47 |

Note. 2PL = 2-parameter logistic model; RMSE = root-mean-squared-error; Statistics calculated from human and RoBERTa's response data and their comparisons are shown. The 2PL model parameters for humans and RoBERTa have been placed on the same scale. RoBERTa's ability was calculated using human and RoBERTa item parameters.

[a]With an outlier removed (item 22 on Figure 3), bias improved to -0.22, and RMSE to 0.73.

[b]Bias and RMSE are not shown as they would be misleading comparisons